\documentclass[11pt,a4paper]{article}
\usepackage[hyperref]{eacl2021}
\usepackage{amsfonts,amssymb}
\usepackage{graphicx}
\usepackage{amsmath,bm}
\usepackage{booktabs}

\usepackage{booktabs}
\usepackage{makecell}
\usepackage{tabularx}
\usepackage{multirow}
\usepackage{arydshln}
\usepackage{tabu}
\usepackage{xcolor}

\usepackage{amsmath}
\usepackage{amssymb}
\usepackage{bm}

\usepackage{ dsfont }

\usepackage{microtype}

\usepackage{enumitem}

\usepackage{graphicx}
\usepackage{caption}

\usepackage{esvect}

\usepackage{microtype}

\aclfinalcopy 

\newcommand{\yulan}[1]{\textcolor{black}{{#1}}}

\newcommand{\gl}[1]{\textcolor{black}{{#1}}}

\newcommand{\gab}[1]{\textcolor{black}{{#1}}}

\title{Adversarial Learning of Poisson Factorisation Model for Gauging Brand Sentiment in User Reviews}
\author{Runcong Zhao, Lin Gui,  Gabriele Pergola,  Yulan He\\
         Department of Computer Science, University of Warwick, UK\\
         \texttt{\{runcong.zhao,lin.gui,gabriele.pergola,yulan.he\}@warwick.ac.uk}
         }

\date{}

\begin{document}
\maketitle
\begin{abstract}

\yulan{In this paper, we propose the Brand-Topic Model (BTM) which aims to detect brand-associated polarity-bearing topics from product reviews. Different from existing models for sentiment-topic extraction which assume topics are grouped under discrete sentiment categories such as `\emph{positive}', `\emph{negative}' and `\emph{neural}', BTM is able to automatically infer real-valued brand-associated sentiment scores and generate fine-grained sentiment-topics in which we can observe continuous changes of words under a certain topic (e.g., `\emph{shaver}' or `\emph{cream}') while its associated sentiment gradually varies from negative to positive. BTM is built on the Poisson factorisation model with the incorporation of adversarial learning. It has been evaluated on a dataset constructed from Amazon reviews. Experimental results show that BTM outperforms a number of competitive baselines in brand ranking, achieving a better balance of topic coherence and uniqueness, and extracting better-separated polarity-bearing topics.}

\end{abstract}

\section{Introduction}
\yulan{Market intelligence aims to gather data from a company's external environment, such as customer surveys, news outlets and social media sites, in order to understand customer feedback to their products and services and to their competitors, for a better decision making of their marketing strategies. Since consumer purchase decisions are heavily influenced by online reviews, it is important to automatically analyse customer reviews for online brand monitoring. Existing sentiment analysis models either classify reviews into discrete polarity categories such as `\emph{positive}', `\emph{negative}' or `\emph{neural}', or perform more fine-grained sentiment analysis, in which aspect-level sentiment label is predicted, though still in the discrete polarity category space. We argue that it is desirable to be able to detect subtle topic changes under continuous sentiment scores. This allows us to identify, for example, whether customers with slightly negative views share similar concerns with those holding strong negative opinions; and what positive aspects are praised by customers the most. In addition, deriving brand-associated sentiment scores in a continuous space makes it easier to generate a ranked list of brands, allowing for easy comparison.} 

\begin{figure}[t]
\centering
\includegraphics[width=1.0\linewidth]{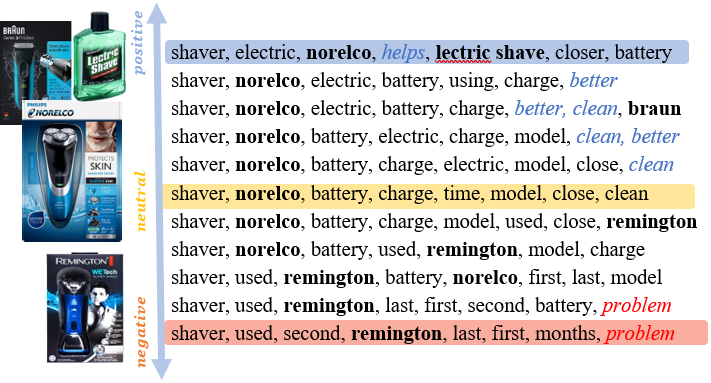}
\caption{Example topic results generated from proposed Brand-Topic Model. We observe a transition of topics with varying topic polarity scores. Besides the change of sentiment-related words (e.g., `\emph{problem}' in negative topics and `\emph{better}' in positive topics), we could also see a change of their associated brands. Users are more positive about \textsc{Braun}, negative about \textsc{Remington}, and have mixed opinions on \textsc{Norelco}.
}
\label{fig:brand-topics}
\end{figure}

\yulan{Existing studies on brand topic detection were largely built on the Latent Dirichlet Allocation (LDA) model \cite{blei2003latent} which assumes that latent topics are shared among competing brands for a certain market. They however are not able to separate positive topics from negative ones. Approaches to polarity-bearing topic detection can only identify topics under discrete polarity categories such as `\emph{positive}' and `\emph{negative}'. We instead assume that each brand is associated with a latent real-valued sentiment score falling into the range of $[-1,1]$ in which $-1$ denotes negative, $0$ being neutral and $1$ positive, and propose a Brand-Topic Model built on the Poisson Factorisation model with adversarial learning. Example outputs generated from BTM are shown in Figure~\ref{fig:brand-topics} in which we can observe a transition of topics with varying topic polarity scores together with their associated brands. }

\yulan{More concretely, in BTM, a document-word count matrix is factorised into a product of two positive matrices, a document-topic matrix and a topic-word matrix. A word count in a document is assumed drawn from a Poisson distribution with its rate parameter defined as a product of a document-specific topic intensity and its word probability under the corresponding topic, summing over all topics. We further assume that each document is associated with a brand-associated sentiment score and a latent topic-word offset value. The occurrence count of a word is then jointly determined by both the brand-associated sentiment score and the topic-word offset value. The intuition behind is that if a word tends to occur in documents with positive polarities, but the brand-associated sentiment score is negative, then the topic-word offset value will have an opposite sign, forcing the occurrence count of such a word to be reduced. Furthermore, for each document, we can sample its word counts from their corresponding Poisson distributions and form a document representation which is subsequently fed into a sentiment classifier to predict its sentiment label. If we reverse the sign of the latent brand-associated sentiment score and sample the word counts again, then the sentiment classifier fed with the resulting document representation should generate an opposite sentiment label.  }


\yulan{Our proposed BTM is partly inspired by the recently developed Text-Based Ideal Point (TBIP) model \cite{tbip} in which the topic-specific word choices are influenced by the ideal points of authors in political debates. However, TBIP is fully unsupervised and when used in customer reviews, it generates topics with mixed polarities. On the contrary, BTM makes use of the document-level sentiment labels and is able to produce better separated polarity-bearing topics. As will be shown in the experiments section, BTM outperforms TBIP on brand ranking, achieving a better balance of topic coherence and topic uniqueness measures. }

\gl{The contributions of the model are three-fold:} 
\begin{itemize}
    \item \gl{We propose a novel model built on Poisson Factorisation with adversarial learning for brand topic analysis which can disentangle the sentiment factor from the semantic latent representations to achieve a flexible and controllable topic generation;}
    \item \gl{We approximate word count sampling from Poisson distributions by the Gumbel-Softmax-based word sampling technique, and construct document representations based on the sampled word counts, which can be fed into a sentiment classifier, allowing for end-to-end learning of the model;}
    \item  \gab{The model, trained with the supervision of review ratings, is able to automatically infer the brand polarity scores from review text only. }
\end{itemize}
\gl{The rest of the paper is organised as follows. Section 2 presents the related work. 
Section 3 describes our proposed Brand-Topic Model. Section 4 and 5  discusses the experimental setup and evaluation results, respectively. Finally, Section 5 concludes the paper and outlines the future research directions.} 



\section{Related Work}

\yulan{Our work is related to the following research:}

\paragraph{Poisson Factorisation Models}
 
\yulan{Poisson factorisation is a class of non-negative matrix factorisation in which a matrix is decomposed into a product of matrices.}  \gl{It has been used in many personalise application such as personalised budgets recommendation \cite{DBLP:conf/ijcai/GuoXSW17}, ranking \cite{DBLP:conf/sdm/KuoCC18}, or content-based social recommendation \cite{DBLP:conf/ijcnn/SuLTZXG19,DBLP:conf/pkdd/SilvaLR17}.}   

\yulan{Poisson factorisation can also be used for topic modelling where a document-word count matrix is factorised into a product of two positive matrices, a document-topic matrix and a topic-word matrix \cite{gan2015scalable,jiang2017topic}. In such a setup, a word count in a document is assumed drawn from a Poisson distribution with its rate parameter defined as a product of a document-specific topic intensity and its word probability under the corresponding topic, summing over all topics. }



\paragraph{Polarity-bearing Topics Models} 

\yulan{Early approaches to polarity-bearing topics extraction were built on LDA in which a word is assumed to be generated from a corpus-wide 
sentiment-topic-word distributions \cite{jst2009}. In order to be able to separate topics bearing different polarities, word prior polarity knowledge needs to be incorporated into model learning. }
\gl{
In recent years, the neural network based topic models have been proposed for many NLP tasks, such as information retrieval \cite{DBLP:conf/kdd/XieDX15}, aspect extraction \cite{DBLP:conf/acl/HeLND17} and sentiment classification \cite{DBLP:conf/coling/HeLND18}. Most of them are built upon Variational Autoencode (VAE) \cite{DBLP:journals/corr/KingmaW13} which constructs a neural network to approximate the topic-word distribution in probabilistic topic models \cite{srivastava2017autoencoding,DBLP:conf/nips/SonderbyRMSW16,DBLP:conf/aaai/BouchacourtTN18}. 
Intuitively, training the VAE-based supervised neural topic models with class labels \cite{DBLP:conf/sigir/ChaidaroonF17,huang2018siamese,9112648} can introduce sentiment information into topic modelling, which may generate better features for sentiment classification. }

\paragraph{Market/Brand Topic Analysis} 

The classic LDA can also be used to analyse market \yulan{segmentation and brand reputation} in various fields such as finance and medicine \citep{aat2018,ftm2009}. For market analysis, the model proposed by \citet{ttm2009} used topic tracking to analyse customers' purchase probabilities and trends without storing historical data for inference \yulan{at the current time step}. Topic analysis can also be combined with additional market information for recommendations. For example, based on user profiles and item topics, \citet{dstr2017} dynamically modelled users' interested items for recommendation. 
\citet{dtm2015} focused on brand topic tracking. They built a dynamic topic model to \yulan{analyse texts and images posted on Twitter and} track competitions in the luxury market among given brands, in which topic words were used to identify recent hot topics in the market (e.g. \emph{Rolex watch}) and brands over topics were used to identify the market share of each brand. 

\paragraph{Adversarial Learning}
\gab{Several studies have explored the application of adversarial learning mechanics to text processing for style transferring \cite{Vineet19}, disentangling representations \cite{Vineet19} and topic modelling \cite{Tomonari18}. }\gab{In particular, \citet{Rui19} has proposed an Adversarial-neural Topic Model (ATM) based on the Generative Adversarial Network (GAN) \cite{Goodfellow14}, that employees an adversarial approach to train a generator network producing word distributions indistinguishable from topic distributions in the training set. 
\citep{wang20} further extended the ATM model with a Bidirectional Adversarial Topic (BAT) model, using a bidirectional adversarial training to incorporate a Dirichlet distribution as prior and exploit the information encoded in word embeddings. 
Similarly, \citep{Hu20} builds on the aforementioned adversarial approach adding cycle-consistent constraints. }

\gab{
Although the previous methods make use of adversarial mechanisms to approximate the posterior distribution of topics, to the best of our knowledge, none of them has so far used adversarial learning to lead the generation of topics based on their sentiment polarity and they do not provide any mechanism for smooth transitions between topics, as introduced in the presented Brand-Topic Model.
}

\section{Brand-Topic Model (BTM)}

\begin{figure*}[h]
\centering
\includegraphics[width=0.95\textwidth]{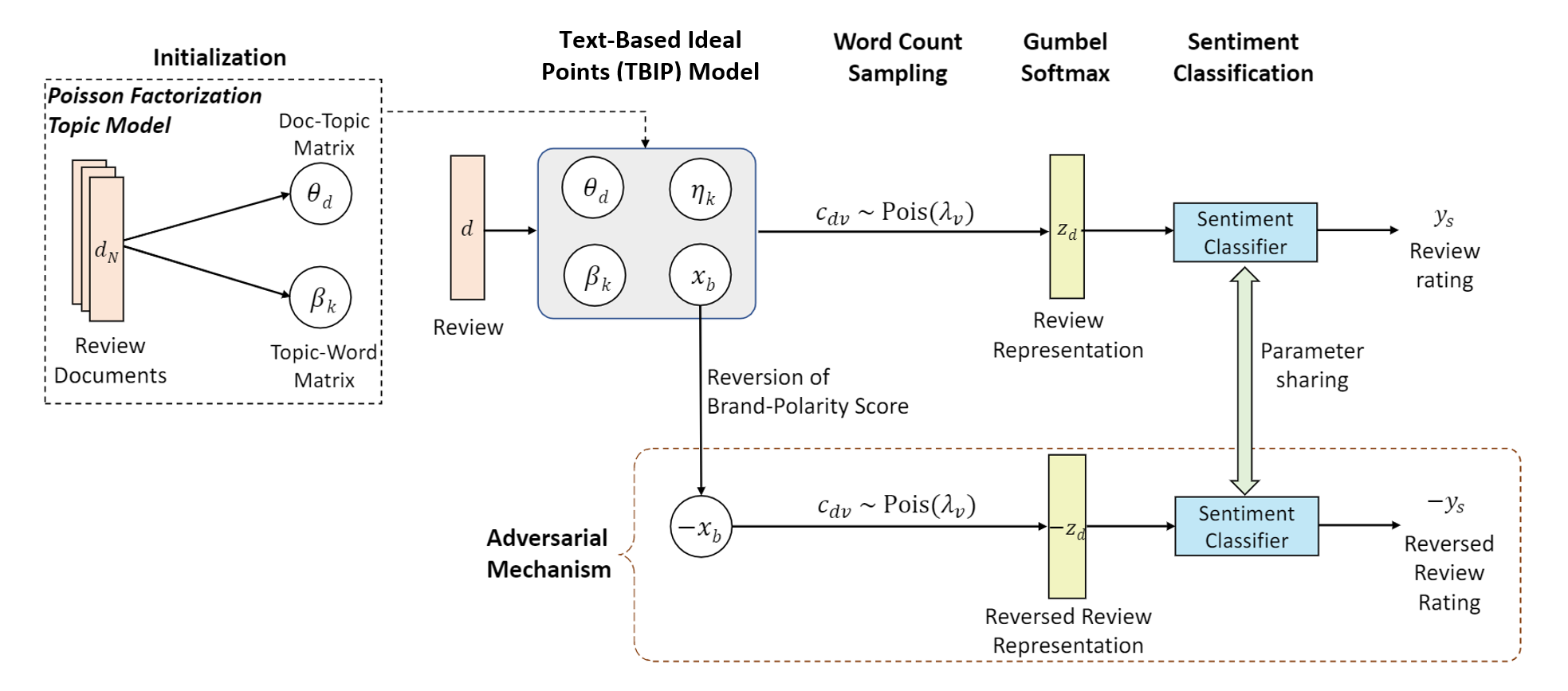}
\caption{The overall architecture of the Brand-Topic Model.}
\label{fig:architecture}
\end{figure*}

We propose a probabilistic model for monitoring the assessment of various brands in the beauty market from Amazon reviews. We extend the Text-Based Ideal Point (TBIP) model with adversarial learning and Gumbel-Softmax to construct document features for sentiment classification. The overall architecture of our proposed BTM is shown in Figure \ref{fig:architecture}. \yulan{In what follows, we will first give a brief introduction of TBIP, followed by the presentation of our proposed BTM}. 

\subsection{Background: Text-Based Ideal Point (TBIP) model}

\gab{TBIP \cite{tbip} is a probabilistic model which aims to quantify political positions (i.e. ideal points) from politicians' speeches and tweets via Poisson factorisation.
In its generative processes, political text is generated from the interactions of several latent variables: the per-document topic intensity $\theta_{dk}$ for $K$ topics and $D$ documents, the $V$-vectors representing the topics $\beta_{kv}$ with vocabulary size $|V|$, the author's ideal point $s$ expressed with a real-valued scalar $x_{s}$ and the ideological topic expressed by a real-valued $V$-vector $\eta_k$.
In particular, the ideological topic $\eta_k$ aligns the neutral topic (e.g. \emph{gun}, \emph{abortion}, etc.) according to the author's ideal point (e.g. \emph{liberal}, \emph{neutral}, \emph{conservative}), thus modifying the prominent words in the original topic (e.g. '\emph{gun violence}', or '\emph{constitutional rights}').
The observed variables are the author $a_d$ for a document $d$, and the word count for a term $v$ in $d$ encoded as $c_{dv}$
}. 

The TBIP model places a Gamma prior on $\bm{\beta}$ and $\bm{\theta}$, which is the assumption inherited from the Poisson factorisation, with $m$, $n$ being hyper-parameters. 
\begin{equation}
    \theta_{dk} \sim \mbox{Gamma}(m,n) \quad
    \beta_{kv} \sim \mbox{Gamma}(m,n) \nonumber
\end{equation}

\noindent \gab{It places instead a normal prior over the ideological topic $\bm{\eta}$ and ideal point $\bm{x}$: }
%
\begin{equation}
    \eta_{kv} \sim \mathcal{N}(0,1) \quad
    x_{s} \sim \mathcal{N}(0,1) \nonumber
\end{equation}
The word count for a term $v$ in $d$, $c_{dv}$, can be modelled with Poisson distribution:
\begin{equation}
    c_{dv} \sim \text{Pois}(\sum_k{\theta_{dk}\beta_{kv}\exp\{x_{a_{d}}\eta_{kv}\})}
\end{equation}

%

\subsection{Brand-Topic Model (BTM)}
\gab{Inspired by the TBIP model, we introduce the Brand-Topic Model by reinterpreting the ideal point $x_{s}$ as brand-polarity score $x_b$ expressing an \textit{ideal feeling} derived from reviews related to a brand, and the ideological topics $\eta_{kv}$ as \textit{opinionated topics}, i.e. polarised topics about brand qualities}.


\gab{Thus, a term count $c_{dv}$ for a product's reviews derives from the hidden variable interactions as $c_{dv} \sim Pois({\lambda}_{dv})$ where:
\begin{equation}
    {\lambda}_{dv} = \sum_k{\theta_{dk}\exp\{\log\beta_{kv} + x_{b_{d}}\eta_{kv}\})}
\end{equation}
}
\noindent with the priors over $\bm{\beta}$, $\bm{\theta}$, $\bm{\eta}$ and $\bm{x}$ initialised according to the TBIP model.

The intuition is that if a word tends to frequently occur in reviews with positive polarities, but the brand-polarity score for the current brand is negative, then the occurrence count of such a word would be reduced since $x_{b_{d}}$ and $\eta_{kv}$ have opposite signs.

\paragraph{Distant Supervision and Adversarial Learning}

\gab{Product reviews might contain opinions about products and more general users' experiences (e.g. delivery service), which are not strictly related to the product itself and could mislead the inference of a reliable brand-polarity score. Therefore, to generate topics which are mainly characterised by product opinions, we provide an additional distant supervision signal via their review ratings. To this aim, we use a sentiment classifier, a simple linear layer, over the generated document representations to infer topics that are discriminative of the review's rating.}

\begin{figure}[!ht]
\centering
\includegraphics[width=0.80\linewidth]{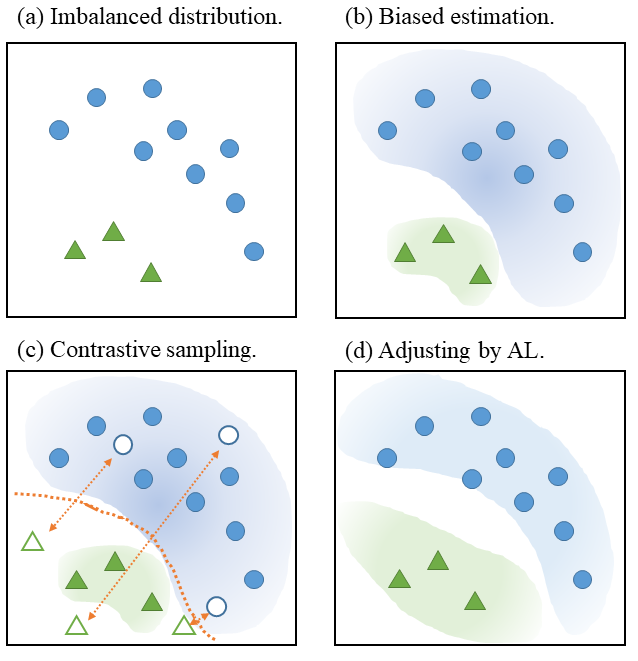}
\caption{\label{fig:adv-learn}
\gl{Process of Adversarial Learning (AL): (a) The imbalanced distribution of different sentiment categories; (b) The biased estimation of distribution from training samples; (c) Contrastive sample generation (white triangles) by reversing the sampling results from biased estimation (white dots); (d) Adjusting the biased estimation of (b) by the contrastive samples.}
}
\end{figure}

\gl{In addition, to deal with the imbalanced distribution in the reviews, we design an adversarial mechanism linking the brand-polarity score to the topics as shown in Figure~\ref{fig:adv-learn}. We contrastively sample adversarial training instances by reversing the original brand-polarity score ($x_b \in [-1,1]$) and generating associated representations. This representation will be fed into the shared sentiment classifier with the original representation to maximise their distance in the latent feature space.}

\paragraph{Gumbel-Softmax for Word Sampling}

As discussed earlier, in order to construct document features for sentiment classification, we need to \yulan{sample word counts} from the Poisson distribution. However, directly sampling word counts from the Poisson distribution is not differentiable. In order to enable back-propagation of gradients, we apply Gumbel-Softmax \citep{gs2017, gs2020}, which is a gradient estimator with 
the reparameterization trick.

\yulan{For a word $v$ in document $d$, its occurrence count,} $c_{dv} \sim \mbox{Pois}(\lambda_{dv})$, is a non-negative random variable with the Poisson rate $\lambda_{dv}$. We can approximate it by \yulan{sampling from the truncated Poisson distribution}, 
$c_{dv_n} \sim \mbox{TruncatedPois}(\lambda_{dv}, n)$, where
\begin{gather}
\pi_k = Pr(c_{dv}=k) = \frac{\lambda_{dv}^ke^{-\lambda_{dv}}}{k!} \nonumber\\
\pi_{n-1} = 1-\sum_k \pi_k \quad \mbox{for}\quad k \in \{0,1,...,n-2\}.\nonumber
\end{gather}

We can then draw samples $z_{dv}$ from the categorical distribution with class probabilities $\pi$ = ($\pi_0$, $\pi_1$, $\cdots$, $\pi_{n-1})$ using:
\begin{gather}
    u_i \sim \mbox{Uniform}(0,1) \quad g_i = -\log(-\log(u_i)) \nonumber \\
    w_i = \mbox{softmax}\big((g_i + \log\pi_i)/\tau\big) \quad 
    z_{dv} = \sum_i w_i c_i \nonumber
\end{gather}
where $\tau$ is a constant referred to as the temperature, $c$ is the outcome vector. By using the average of weighted word account, the process is now differentiable and we use the sampled \yulan{word counts to form the document representation and feed it as an input to the sentiment classifier.} 

\paragraph{Objective Function}

Our final objective function consists of three parts, including the Poisson factorisation model, the sentiment classification loss, and the reversed sentiment classification loss (for adversarial learning). For the Poisson factorisation modelling part, mean-field variational inference is used to approximate posterior distribution \citep{vmgm1999, gmefvi2008, virs2017}.
\begin{equation}
    q_{\phi}(\theta,\beta,\eta,x) = \prod_{d,k.b}q(\theta_d)q(\beta_k)q(\eta_k)q(x_b)
\end{equation}
For optimisation, to minimise the approximation of $q_{\phi}(\theta,\beta,\eta,x)$ and the posterior, equivalently we maximise the evidence lower bound (ELBO): 
\begin{equation}
\begin{split}
    ELBO = \mathbb{E}_{q_{\phi}}[log\ p(\theta,\beta,\eta,x)] +  \\
    log\ p(y|\theta,\beta,\eta,x) -  log\ q_{\phi}(\theta,\beta,\eta,x)]
\end{split}
\end{equation}
The Poisson factorization model is pre-trained by applying the algorithm in \citet{gan2015scalable}, which is then used to initialise the varational parameters of $\theta_d$ and $\beta_k$. 
Our final objective function is:

\begin{equation}
    Loss = ELBO + \lambda(L_s + L_a)
\end{equation}

\noindent where $L_s$ and $L_a$ are the cross entropy loss of sentiment classification for sampled documents and reversed sampled documents, respectively, and $\lambda$ is the weight to balance the two parts of loss, which is set to be 100 in our experiments.




\section{Experimental Setup}

\paragraph{Datasets}
We \yulan{construct our dataset by retrieving reviews in the Beauty category from the} Amazon review corpus\footnote{ \url{http://jmcauley.ucsd.edu/data/amazon/}} \citep{ar2016}. 
\yulan{Each review is accompanied with the rating score (between 1 and 5), reviewer name and the product meta-data such as product ID, description, brand and image. We use the product meta-data} 
to relate a product with its associated brand. \yulan{By only selecting} 
brands with relatively more and balanced reviews, \yulan{our final dataset contains a total of} 78,322 reviews from 45 brands. 
Reviews with the rating score of 1 and 2 are grouped as negative reviews; those with the score of 3 are neutral reviews; and the remaining are positive reviews. \yulan{The statistics of our dataset is shown in Table \ref{tab:dataset_statistics}\footnote{The detailed rating score distributions of brands and their average rating are shown in Table \ref{append:brad_statistics} in the Appendix.}.} We can observe that our data is highly imbalanced, with the positive reviews far more than negative and neutral reviews. 

\gab{
\begin{table}[!ht]
\centering
\resizebox{\columnwidth}{!}{
\begin{tabular}{@{}lr@{}}
    \toprule
    \textbf{Dataset}  & \textbf{Amazon-Beauty Reviews} \\
    \midrule
    Documents per classes   &   \\
    $\ \ \ $ Neg / Neu / Pos   & 9,545 / 5,578 / 63,199  \\
    Brands & 45 \\
    Total \#Documents & 78,322  \\
    Avg. Document Length   & 9.7  \\
    Vocabulary size   & $\sim 5000$ \\
    \bottomrule
\end{tabular}
}
\caption{Dataset statistics of reviews within the Amazon dataset under the \textit{Beauty} category.}
\label{tab:dataset_statistics}
\end{table}
}



\paragraph{Baselines}

\yulan{We compare the performance of our model with the following baselines:}
\begin{itemize}
    \item Joint Sentiment-Topic (JST) model \citep{jst2009}, built on LDA, can extract polarity-bearing topics from text provided that it is supplied with the word prior sentiment knowledge. In our experiments, the MPQA subjectivity lexicon\footnote{\url{https://mpqa.cs.pitt.edu/lexicons/}} is used to derive the word prior sentiment information.
    \item \textsc{Scholar} \citep{scholar2018}, a neural topic model built on VAE. It allows the incorporation of meta-information such as document class labels into the model for training, essentially turning it into a supervised topic model.
    \item Text-Based Ideal Point (TBIP) model, an unsupervised Poisson factorisation model which can infer latent brand sentiment scores.
\end{itemize}


\paragraph{Parameter setting}

\gl{Since documents are represented as the bag-of-words which result in the loss of word ordering or structural linguistics information, frequent bigrams and trigrams 
such as `\emph{without doubt}', `\emph{stopped working}', are also used as features for document representation construction. Tokens, i.e., $n$-grams ($n=\{1,2,3\}$), occurred less than twice are filtered.} 
In our experiments, we set aside 10\% reviews (7,826 reviews) as the test set and the remaining (70,436 reviews) as the training set. For hyperparameters, we set the batch size to 1,024, the maximum training steps to 50,000, the topic number to 30, the temperature in the Gumbel-Softmax equation in Section 3.2 to 1. 
Since our dataset is highly imbalanced, we balance data in each mini-batch by oversampling. For a fair comparison, we report two sets of results from the baseline models, one trained from the original data, the other trained from the balanced training data by oversampling negative reviews. The latter results in an increased training set consisting of 113,730 reviews.



\section{Experimental Results}

In this section, we \yulan{will present the experimental results in comparison with the baseline models in brand ranking, topic coherence and uniqueness measures, and also present the qualitative evaluation of the topic extraction results. We will further discuss the limitations of our model and outline future directions.}

\subsection{Comparison with Existing Models}
\begin{table}[h]
\centering
\begin{tabular}{lcccc} 
\toprule
\multirow{2}{*}{\textbf{Model }} & \multicolumn{2}{c}{\textbf{Spearman's }} & \multicolumn{2}{c}{\textbf{Kendall's tau} }  \\
                                 & corr            & p-val                  & corr            & p-val                      \\ 
\midrule
JST               & 0.241             & 0.111   & 0.180                    & 0.082   \\ 
JST*                    & 0.395           & 0.007                & 0.281           & 0.007                      \\
\textsc{Scholar}               & -0.140            & 0.358   & -0.103                   & 0.318   \\ 
\textsc{Scholar*}   & 0.050             & 0.743   & 0.046                   & 0.653   \\ 
TBIP                   & 0.361           & 0.016                  & 0.264           & 0.012                      \\
BTM                    & \textbf{0.486}  & 0.001                  & \textbf{0.352}  & 0.001                      \\
\bottomrule
\end{tabular}
\caption{Brand ranking results generated by various models based on the test set. We report the correlation coefficients \textit{corr} and its associated two-sided $p$-values for both Spearman's correlations and Kendall's tau. * indicates models trained on balanced training data.}
\label{tab:ranking}
\end{table}

\paragraph{Brand Ranking}


\yulan{We report in Table \ref{tab:ranking} the brand ranking results generated by various models on the test set. The two commonly used evaluation metrics for ranking tasks, Spearman's correlations and Kendall’s Tau, are used here. They penalise inversions equally across the ranked list. Both TBIP and BTM can infer each brand's associated polarity score automatically which can be used for ranking.} 
For both JST and \textsc{Scholar}, \yulan{we derive the polarity score of a brand by aggregating the sentiment probabilities of its associated review documents and then normalising over the total number of brand-related reviews. } 
\yulan{It can be observed} from Table~\ref{tab:ranking} that \yulan{JST outperforms both \textsc{Scholar} and TBIP.} \gl{Balancing the distributions of sentiment classes improves the performance of JST and \textsc{Scholar}.  
Overall, BTM gives the best results, showing the effectiveness of adversarial learning.} 

\paragraph{Topic Coherence and Uniqueness}


Here we choose the top 10 words for each topics \gl{to calculate the context-vector-based topic coherence scores \cite{DBLP:conf/wsdm/RoderBH15}}. 
\yulan{In the topics generated by TBIP and BTM, we can vary the topic polarity scores to generate positive, negative and neutral subtopics as shown in Table \ref{tab:topic-example}. We would like to achieve high topic coherence, but at the same time maintain a good level of topic uniqueness across the sentiment subtopics since they express different polarities.} Therefore, we additionally consider the topic uniqueness \citep{tmwa2019} to measure word redundancy \yulan{among sentiment subtopics}, 
$TU = \frac{1}{LK} \sum_{l=1}^{K}\sum_{l=1}^{L}{\frac{1}{cnt(l,k)}}$, where $cnt(l,k)$ denotes the number of times word $l$ appear across \emph{positive}, \emph{neutral} and \emph{negative} topics under the same topic number $k$.
We can see from Table~\ref{tab:coherence} that \yulan{both TBIP and BTM achieve higher coherence scores compared to JST and \textsc{Scholar}. TBIP slightly outperforms BTM on topic coherence, but has a lower topic uniqueness score. As will be shown in Table \ref{tab:topic-example}, topics extracted by TBIP contain words significantly overlapped with each other among sentiment subtopics. \textsc{Scholar} gives the highest topic uniqueness score. However, it cannot separate topics with different polarities. Overall, our proposed BTM achieves the best balance between topic coherence and topic uniqueness}.  

\begin{table}[htb]
 \centering
\begin{tabular} {lccc} 
\toprule
\makecell[l]{Model} & \makecell[c]{Topic\\Coherence} &  \makecell[c]{Topic\\Uniqueness} \\ \midrule 
JST            & 0.1423                   & 0.7699              \\
JST*            & 0.1317                   & 0.7217                  \\
\textsc{Scholar}        & 0.1287                   & \textbf{0.9640}           \\
\textsc{Scholar}*        & 0.1196                   & 0.9256        \\
TBIP           & \textbf{0.1525}          & 0.8647                  \\
BTM            & 0.1407                   & 0.9033                  \\
\bottomrule
\end{tabular}
\caption{Topic coherence/uniqueness measures of results generated by various models. }
\label{tab:coherence}
\end{table}

\begin{table*}[ht]
\centering
\begin{tabular}{lll}
\toprule
\textbf{Topic}                  & \textbf{Sentiment} & \multicolumn{1}{c}{\multirow{2}{*}{\textbf{Top Words}}}                                                        \\
\textbf{Label}                     & \textbf{Topics}    & \multicolumn{1}{c}{}                                                                                  \\  \hline
\multicolumn{3}{c}{\textbf{BTM}}                                                                                                                        \\ \hline  
\multirow{3}{*}{Brush}     & Positive  & brushes, \textcolor{black}{cheap}, came, pay, \textcolor{blue}{\textit{pretty}}, brush, \textcolor{blue}{\textit{okay}}, case, glue, \textcolor{blue}{\textit{soft}} \\
                           & Neutral   & \textcolor{black}{cheap}, feel, set, buy, \textcolor{red}{\textit{cheaply made}}, feels, made, worth, spend, bucks                                 \\
                           & Negative  & plastic, made, \textcolor{black}{cheap}, parts, feels, \textcolor{red}{\textit{flimsy}}, money, \textcolor{red}{\textit{break}}, metal, bucks                                \\ \hline
                           
\multirow{3}{*}{Oral Care} & Positive  & teeth, taste, mouth, strips, crest, mouthwash, tongue, using, \textcolor{blue}{\textit{white}}, rinse                            \\
                           & Neutral   & teeth, \textcolor{red}{\textit{pain}}, mouth, strips, using, taste, used, crest, mouthwash, \textcolor{blue}{\textit{white}}                               \\
                           & Negative  & \textcolor{red}{\textit{pain}}, \textcolor{red}{\textit{issues}}, causing, teeth, caused, removing, wore, \textcolor{red}{\textit{burn}}, little, cause                             \\ \hline
                           

\multirow{3}{*}{Duration} & Positive  & stay, pillow, \textcolor{blue}{\textit{comfortable}}, string, tub, mirror, stick, back, months                            \\
                           & Neutral   & months, year, \textcolor{blue}{\textit{lasted}}, \textcolor{red}{\textit{stopped working}}, \textcolor{red}{\textit{sorry}}, n, worked, working, u, last                              \\
                           & Negative  & months, year, last, \textcolor{blue}{\textit{lasted}}, battery, warranty, \textcolor{red}{\textit{stopped working}}, \textcolor{red}{\textit{died}}, \textcolor{red}{\textit{less}}                             \\ \hline 
                           
\multicolumn{3}{c}{\textbf{TBIP}}                                                                                                                       \\ \hline 
\multirow{3}{*}{Brush}      & Positive  & love, \textcolor{blue}{\textit{favorite}}, products, \textcolor{blue}{\textit{definitely recommend}}, forever, carry, \textcolor{black}{brushes} \\
                           & Neutral   & love, brushes, \textcolor{blue}{\textit{cute}}, \textcolor{blue}{\textit{favorite}}, \textcolor{blue}{\textit{definitely recommend}}, soft, \textcolor{red}{\textit{cheap}}             \\
                           & Negative  & love, brushes, cute, \textcolor{blue}{\textit{soft}}, \textcolor{red}{\textit{cheap}}, set, case, quality price, buy, bag                                  \\ \hline
                           
\multirow{3}{*}{Oral Care} & Positive  & teeth, strips, crest, mouth, mouthwash, taste, \textcolor{blue}{\textit{white}}, \textcolor{blue}{\textit{whitening}}, sensitivity              \\
                           & Neutral   & teeth, strips, mouth, crest, taste, work, \textcolor{red}{\textit{pain}}, using, \textcolor{blue}{\textit{white}}, mouthwash                               \\
                           & Negative  & teeth, strips, mouth, crest, taste, work, \textcolor{red}{\textit{pain}}, using, \textcolor{blue}{\textit{white}}, mouthwash                               \\ \hline
                           

\multirow{3}{*}{Duration} & Positive  & great, \textcolor{blue}{\textit{love shampoo}}, \textcolor{blue}{\textit{great price}}, \textcolor{blue}{\textit{great product}}, \textcolor{blue}{\textit{lasts long time}}          \\
                           & Neutral   & \textcolor{blue}{\textit{great}}, \textcolor{blue}{\textit{great price}}, \textcolor{blue}{\textit{lasts long time}}, \textcolor{blue}{\textit{great product}}, price, \textcolor{blue}{\textit{works expected}}                          \\
                           & Negative  & quality, \textcolor{blue}{\textit{great}}, \textcolor{blue}{\textit{fast shipping}}, \textcolor{blue}{\textit{great price}}, \textcolor{blue}{\textit{low price}}, price quality, hoped \\ 
\bottomrule
\end{tabular}
\caption{\label{tab:topic-example}
Example topics generated by BTM and TBIP on Amazon reviews. The topic labels are assigned by manual inspection. Positive words are highlighted with the blue colour, while negative words are marked with the red colour. BTM generates better-separated sentiment topics compared to TBIP.
}
\end{table*}

\subsection{Example Topics Extracted from Amazon Reviews}

\yulan{We illustrate some representative topics generated by TBIP and BTM in Table~\ref{tab:topic-example}. It is worth noting that we can generate a smooth transition of topics by varying the topic polarity score gradually as shown in Figure~\ref{fig:brand-topics}. Due to space limit, we only show topics when the topic polarity score takes the value of $-1$ (\emph{negative}), $0$ (\emph{neutral}) and $1$ (\emph{positive}). It can be observed that TBIP fails to separate subtopics bearing different sentiments. For example, all the subtopics under `Duration' express a positive polarity. On the contrary, BTM shows a better-separated sentiment subtopics. For `Duration', we see positive words such as `\emph{comfortable}' under the positive subtopic, and words such as `\emph{stopped working}' clearly expressing negative sentiment under the negative subtopic. Moreover, top words under different sentiment subtopics largely overlapped with each other for TBIP. But we observe a more varied vocabulary in the sentiment subtopics for BTM. }

\yulan{TBIP was originally proposed to deal with political speeches in which speakers holding different ideal points tend to use different words to express their stance on the same topic. This is however not the case in Amazon reviews where the same word could appear in both positive and negative reviews.}
For example, `\emph{cheap}' for lower-priced products could convey a positive polarity to \yulan{express value for money}, 
but it could also bear a negative polarity implying a poor quality. \yulan{As such, it is difficult for TBIP to separate words under different polarity-bearing topics. On the contrary, with the incorporation of adversarial learning, our proposed BTM is able to extract different set of words co-occurred with `\emph{cheap}' under topics with different polarities, thus accurately capturing the contextual polarity of the word `\emph{cheap}'. For example, `\emph{cheap}' appears in both positive and negative subtopics for `Brush' in Table~\ref{tab:topic-example}. But we can find other co-occurred words such as `\emph{pretty}' and `\emph{soft}' under the positive subtopic, and `\emph{plastic}' and `\emph{flimsy}' under the negative subtopic, which help to infer the contextual polarity of `\emph{cheap}'. } 

\yulan{TBIP also appears to have a difficulty in dealing with highly imbalanced data. In our constructed dataset, positive reviews significantly outnumber both negative and neutral ones. In many sentiment subtopics extracted by TBIP, all of them convey a positive polarity. One example is the `Duration' topic under TBIP, where words such as `\emph{great}', `\emph{great price}' appear in all positive, negative and neutral topics. With the incorporation of supervised signals such as the document-level sentiment labels, our proposed BTM is able to derive better separated polarised topics.} 

As an example shown in 
Figure~\ref{fig:brand-topics}, if we vary the polarity score of a topic from $-1$ to $1$, we observe a smooth transition of its associated topic words, gradually moving from negative to positive. Under the topic (\emph{shaver}) shown in this figure, four brand names appeared: \textsc{Remington}, \textsc{Norelco}, \textsc{Braun} and \textsc{Lectric Shave}. The first three brands can be found in our dataset. \textsc{Remington} appears in the negative side and it indeed has the lowest review score among these 3 brands; \textsc{Norelco} appears most and it is indeed a popular brand with mixed reviews; and \textsc{Braun} gets the highest score in these 3 brands, which is also consistent with the observations in our data. Another interesting finding is the brand \textsc{Lectric Shave}, which is not one of the brands we have in the dataset. But we could predict from the results that it is a product with relatively good reviews.


\subsection{Limitations and Future work}
Our model requires the use of a vanilla Poisson factorisation model to initialise the topic distributions before applying the adversarial learning mechanism of BTM to perform a further split of topics based on varying polarities. Essentially topics generated by a vanilla Poisson factorisation model can be considered as parent topics, while polarity-bearing subtopics generated by BTM can be considered as child topics. Ideally, we would like the parent topics to be either neutral or carrying a mixed sentiment which would facilitate the learning of polarised sub-topics better. In cases when parent topics carry either strongly positive or strongly negative sentiment signals, BTM would fail to produce polarity-varying subtopics. 
\gab{One possible way is to employ earlier filtering of topics with strong polarities.} For example, \gab{topic labeling \cite{bhatia16} could be employed to obtain a rough estimate of initial topic polarities; these labels would be in turn used for filtering out topics carrying strong sentiment polarities.}



\gab{Although the adversarial mechanism tends to be robust with respect to class imbalance, the  disproportion of available reviews with different polarities could hinder the model performance. One promising approach suitable for the BTM adversarial mechanism would consist in decoupling the representation learning and the classification, as suggested in \citet{kang19}, preserving the original data distribution used by the model to estimate the brand score.}




\section{Conclusion}
In this paper, we presented the Brand-Topic Model, 
a probabilistic model \gab{which is able to generate polarity-bearing topics of commercial brands. Compared to other topic models, BMT infers real-valued brand-associated sentiment scores and extracts fine-grained sentiment-topics which vary smoothly in a continuous range of polarity scores. 
It builds on the Poisson factorisation model, combining it with an adversarial learning mechanism to induce better-separated polarity-bearing topics. Experimental evaluation on Amazon reviews against several baselines shows an overall improvement of topic quality in terms of coherence, uniqueness and separation of polarised topics.
} 


\section*{Acknowledgements}

This work is funded by the EPSRC (grant no. EP/T017112/1, EP/V048597/1). YH is supported by a Turing AI Fellowship funded by the UK Research and Innovation (UKRI) (grant no. EP/V020579/1).

\bibliography{anthology,eacl2021}
\bibliographystyle{acl_natbib}

\newpage
\onecolumn
\appendix

\section{Appendix}
\label{sec:appendix}

\setcounter{table}{0}
\renewcommand{\thetable}{\Alph{section}\arabic{table}}

\begin{table*}[!h]
\centering
\resizebox{0.98\textwidth}{!}{
\begin{tabular}{|c|c|c|c|c|c|c|c|}
\hline
\multirow{2}{*}{\textbf{Brand}} & \multirow{2}{*}{\textbf{Average Rating}\textbf{}} & \multirow{2}{*}{\textbf{Number of Reviews}} & \multicolumn{5}{c|}{\textbf{Distribution of Ratings}} \\ \cline{4-8} 
                       &                                &                                    & \textbf{1}      & \textbf{2}      & \textbf{3}      & \textbf{4}      & \textbf{5}      \\ \hline
General                & 3.478                          & 1103                               & 236    & 89     & 144    & 180    & 454    \\ \hline
VAGA                   & 3.492                          & 1057                               & 209    & 116    & 133    & 144    & 455    \\ \hline
Remington              & 3.609                          & 1211                               & 193    & 111    & 149    & 282    & 476    \\ \hline
Hittime                & 3.611                          & 815                                & 143    & 62     & 110    & 154    & 346    \\ \hline
Crest                  & 3.637                          & 1744                               & 352    & 96     & 159    & 363    & 774    \\ \hline
ArtNaturals            & 3.714                          & 767                                & 138    & 54     & 65     & 143    & 368    \\ \hline
Urban Spa              & 3.802                          & 1279                               & 118    & 105    & 211    & 323    & 522    \\ \hline
GiGi                   & 3.811                          & 1047                               & 151    & 79     & 110    & 184    & 523    \\ \hline
Helen Of Troy          & 3.865                          & 3386                               & 463    & 20     & 325    & 472    & 1836   \\ \hline
Super Sunnies          & 3.929                          & 1205                               & 166    & 64     & 126    & 193    & 666    \\ \hline
e.l.f                  & 3.966                          & 1218                               & 117    & 85     & 148    & 241    & 627    \\ \hline
AXE PW                 & 4.002                          & 834                                & 85     & 71     & 55     & 169    & 454    \\ \hline
Fiery Youth            & 4.005                          & 2177                               & 208    & 146    & 257    & 381    & 1185   \\ \hline
Philips Norelco        & 4.034                          & 12427                              & 1067   & 818    & 1155   & 2975   & 6412   \\ \hline
Panasonic              & 4.048                          & 2473                               & 276    & 158    & 179    & 419    & 1441   \\ \hline
SilcSkin               & 4.051                          & 710                                & 69     & 49     & 58     & 135    & 399    \\ \hline
Rimmel                 & 4.122                          & 911                                & 67     & 58     & 99     & 160    & 527    \\ \hline
Avalon Organics        & 4.147                          & 1066                               & 111    & 52     & 82     & 145    & 676    \\ \hline
L'Oreal Paris          & 4.238                          & 973                                & 88     & 40     & 72     & 136    & 651    \\ \hline
OZ Naturals            & 4.245                          & 973                                & 79     & 43     & 74     & 142    & 635    \\ \hline
Andalou Naturals       & 4.302                          & 1033                               & 58     & 57     & 83     & 152    & 683    \\ \hline
Avalon                 & 4.304                          & 1344                               & 132    & 62     & 57     & 108    & 985    \\ \hline
TIGI                   & 4.319                          & 712                                & 53     & 32     & 42     & 93     & 492    \\ \hline
Neutrogena             & 4.331                          & 1200                               & 91     & 55     & 66     & 142    & 846    \\ \hline
Dr. Woods              & 4.345                          & 911                                & 60     & 42     & 74     & 83     & 652    \\ \hline
Gillette               & 4.361                          & 2576                               & 115    & 94     & 174    & 555    & 1638   \\ \hline
Jubujub                & 4.367                          & 1328                               & 53     & 42     & 132    & 238    & 863    \\ \hline
Williams               & 4.380                          & 1887                               & 85     & 65     & 144    & 347    & 1246   \\ \hline
Braun                  & 4.382                          & 2636                               & 163    & 85     & 147    & 429    & 1812   \\ \hline
Italia-Deluxe          & 4.385                          & 1964                               & 96     & 73     & 134    & 336    & 1325   \\ \hline
Booty Magic            & 4.488                          & 728                                & 28     & 7      & 48     & 144    & 501    \\ \hline
Greenvida              & 4.520                          & 1102                               & 55     & 33     & 51     & 108    & 855    \\ \hline
Catrice                & 4.527                          & 990                                & 49     & 35     & 34     & 99     & 773    \\ \hline
NARS                   & 4.535                          & 1719                               & 60     & 36     & 107    & 237    & 1279   \\ \hline
Astra                  & 4.556                          & 4578                               & 155    & 121    & 220    & 608    & 3474   \\ \hline
Heritage Products      & 4.577                          & 837                                & 25     & 18     & 52     & 96     & 646    \\ \hline
Poppy Austin           & 4.603                          & 1079                               & 36     & 31     & 38     & 115    & 859    \\ \hline
Aquaphor               & 4.633                          & 2882                               & 100    & 58     & 106    & 272    & 2346   \\ \hline
KENT                   & 4.636                          & 752                                & 23     & 8      & 42     & 74     & 605    \\ \hline
Perfecto               & 4.801                          & 4862                               & 44     & 36     & 81     & 523    & 4178   \\ \hline
Citre Shine            & 4.815                          & 713                                & 17     & 5      & 3      & 43     & 645    \\ \hline
Bath $\&$ Body Works      & 4.819                          & 2525                               & 60     & 27     & 20     & 95     & 2323   \\ \hline
Bonne Bell             & 4.840                          & 1010                               & 22     & 9      & 6      & 35     & 938    \\ \hline
Yardley                & 4.923                          & 788                                & 3      & 4      & 3      & 31     & 747    \\ \hline
Fruits $\&$ Passion       & 4.932                          & 776                                & 3      & 2      & 3      & 29     & 739    \\ \hline
\textbf{Overall}                & 4.259                          & 78322                              & 5922   & 3623   & 5578   & 12322  & 50877  \\ \hline
\end{tabular}
}
\caption{Brand Statistics. The table shows the average rating score, the total number of associated reviews, and the distribution of the number of reviews for ratings ranging between 1 star to 5 stars, for each of the 45 brands.}
\label{append:brad_statistics}
\end{table*}

\end{document}